\def\year{2020}\relax
\documentclass[letterpaper]{article} 
\usepackage{aaai20}  
\usepackage{times}  
\usepackage{latexsym}
\usepackage{algorithm}      
\usepackage{algpseudocode}
\usepackage{helvet} 
\usepackage{courier}  
\usepackage[hyphens]{url}  
\usepackage{graphicx} 
\usepackage{subfigure}
\usepackage{amsfonts}
\usepackage{multirow}
\usepackage{makecell}
\usepackage{url}

\title{Multi-Zone Unit for Recurrent Neural Networks}
\author{
Fandong Meng\textsuperscript{1},
Jinchao Zhang\textsuperscript{1},
Yang Liu\textsuperscript{2}
and Jie Zhou\textsuperscript{1} \\
\textsuperscript{1}WeChat AI - Pattern Recognition Center  Tencent Inc., China \\
\textsuperscript{2}Department of Computer Science and Technology, Tsinghua University, Beijing, China \\
\{fandongmeng, dayerzhang, withtomzhou\}@tencent.com \\
liuyang2011@tsinghua.edu.cn
}

\begin{document}

\maketitle

\begin{abstract}
Recurrent neural networks (RNNs) have been widely used to deal with sequence learning problems. The input-dependent transition function, which folds new observations into hidden states to sequentially construct fixed-length representations of arbitrary-length sequences, plays a critical role in RNNs. Based on single space composition, transition functions in existing RNNs often have difficulty in capturing complicated long-range dependencies. In this paper, we introduce a new {\bf M}ulti-{\bf z}one {\bf U}nit (MZU) for RNNs. The key idea is to design a transition function that is capable of modeling multiple space composition. The MZU consists of three components: zone generation, zone composition, and zone aggregation. Experimental results on multiple datasets of the character-level language modeling task and the aspect-based sentiment analysis task demonstrate the superiority of the MZU.
\end{abstract}

\section{Introduction}
Processing sequential data of variable length is a major challenge in the
field of natural language processing (NLP). Recurrent Neural Networks (RNNs), Long Short-term Memories (LSTMs)~\cite{lstm1997} and Gated Recurrent Units (GRUs)~\cite{ChoEMNLP} in particular, have recently become one of the most popular tools to approach sequence learning tasks, such as 
handwriting recognition~\cite{graves2013generating}, 
sentiment classification~\cite{Tang:16a,Wang:16}, 
sequence labeling~\cite{huang2015bidirectional,char_LSTM_LM_sequence_labeling}, 
language modeling~\cite{ha2016hypernetworks,zoph2016neural,chung2017hierarchical,mujika2017fast,li2018independently}
and machine translation~\cite{acl2018Chen,meng2018dtmt}. 

RNNs sequentially construct fixed-length representations of arbitrary-length sequences by folding new observations into their hidden states using an input-dependent transition operator. The hidden state $\mathbf{h}_t$ is updated recursively using the previous hidden state $\mathbf{h}_{t-1}$ and the current input $\mathbf{x}_t$ as $\mathbf{h}_t=F(\mathbf{h}_{t-1}, \mathbf{x}_t)$, where $F$ is a differentiable function with learnable parameters, such as multiplying its inputs by a matrix and squashing the result with a non-linear function for a vanilla RNN. The state transition between consecutive hidden states adds a new input to the summary of previous ones. However, this procedure of constructing a new summary from the combination of the previous one and the new input in conventional RNNs only bases on a single space composition, which often has difficulty in capturing complicated long-range dependencies, to allow the hidden state to rapidly adapt to quickly changing modes of the input while still preserving a useful summary of the past. 

Recent studies in neural machine translation show that it is beneficial to linearly project one single-space representation into multi-space representations and then capture useful information from these different representation spaces, e.g., the multi-head attention~\cite{VaswaniEtal2017,acl2018Chen,meng2018dtmt}, which also has been verified effective for capturing multiple semantic aspects from the user utterance in generative dialogue systems~\cite{tao2018get}. Although some effective RNN extensions, e.g., GRU and LSTM, have been proposed with gating units controlling the information flow. The gating units are also generated by such kind of function ($F$) based on single space composition, which is not expressive enough to capture potentially complex information for controlling the information flow.

In this paper, we thus boost the $F$ function through modeling multiple space composition, and propose a {\bf M}ulti-{\bf z}one {\bf U}nit for Recurrent Neural Networks, named MZU. The MZU consists of three components: zone generation, zone composition, and zone aggregation. At each time step, the MZU projects the previous hidden state and the current input into multiple zones (zone generation), then conducts full interactions and compositions on these zones (zone composition), and finally aggregates them to the final representation (zone aggregation). In particular, we propose three effective approaches for the zone composition to verify the MZU. 
The first model exploits self-attention~\cite{VaswaniEtal2017} to draw global dependencies on zones. The second model utilizes graph convolutional network~\cite{kipf2016variational,kipf2016semi} to perform neighborhood mixing on zones by leveraging the graph connectivity structure as a filter. The third model uses the dynamic routing algorithm~\cite{sabour2017dynamic} to model part-whole relationships on zones. To further enhance the MZU, we propose an effective regularization objective to promote the diversity in multiple zones. The architecture of MZU is generic as standard RNNs, thus can be extended to deep MZUs.

We evaluate the MZU on 1) the challenging character-level language modeling task with two standard datasets, namely the well known Penn Treebank~\cite{marcus1993building} and the larger Wikipedia dataset (text8)~\cite{mahoney2011large}; and 2) the aspect-based sentiment analysis (ABSA) task with two datasets of the SemEval 2014 Task 4~\cite{Pontiki:14} in different domains. Experimental results on both tasks demonstrate the superiority of the MZU to previous competitive RNNs and its generalizability across tasks. 

Our main contributions are three-fold:
\begin{itemize}
\item We propose a new and generic Multi-zone Unit for RNNs, along with three effective approaches for the zone composition.
\item To further enhance the MZU, we propose an effective regularization objective to promote the diversity of multiple zones.
\item We provide empirical and visualization analyses to reveal advantages of the MZU, e.g., capturing richer linguistic information and better long sequence processing.
\end{itemize}

\section{Multi-Zone Unit for RNNs}
\subsection{Overview}
MZU maintains a hidden state $\mathbf{h}_{t}$ to summarize past inputs at each time step $t$, updated as:
\begin{eqnarray}
\mathbf{h}_{t} = (1 - \mathbf{g}_t) \odot \mathbf{h}_{t-1} + \mathbf{g}_t \odot \mathbf{\widetilde{h}}_{t}
\end{eqnarray} 
where $\odot$ is an element-wise product, $\mathbf{g}_t$ is the gate, to control the flow of information in the previous hidden state $\mathbf{h}_{t-1}$ and the candidate activation $\mathbf{\widetilde{h}}_{t}$ at the current time step, which is computed as:
\begin{eqnarray}
\mathbf{\widetilde{h}}_{t} = \tanh(M_h(\mathbf{x}_t, \mathbf{h}_{t-1}))  \label{activate}
\end{eqnarray}
where $\mathbf{x}_t$ is the input embedding at time $t$, and the gate $\mathbf{g}_t$ is computed as:
\begin{eqnarray}
\mathbf{\mathbf{g}}_{t} = \sigma(M_g(\mathbf{x}_t, \mathbf{h}_{t-1})) \label{gate}
\end{eqnarray}
where $M_h$ and $M_g$ are multi-zone transformation functions, designed for modeling multiple space composition, and their formal definitions will be described in the next section. 

MZU extends the standard RNN by replacing the linear transformation (i.e., the $F$ function) on $[\mathbf{x}_t, \mathbf{h}_{t-1}]$ with multi-zone transformation functions, i.e. the $M_h$ and the $M_g$ in Eq.(~\ref{activate}) and~(\ref{gate}), as well as maintains a simplified gating mechanism to control the information flow. We can also design more complex gating mechanisms as GRU and LSTM to further enhance MZU.

\begin{figure}[t!]
\begin{center}
      \includegraphics[width=0.45\textwidth]{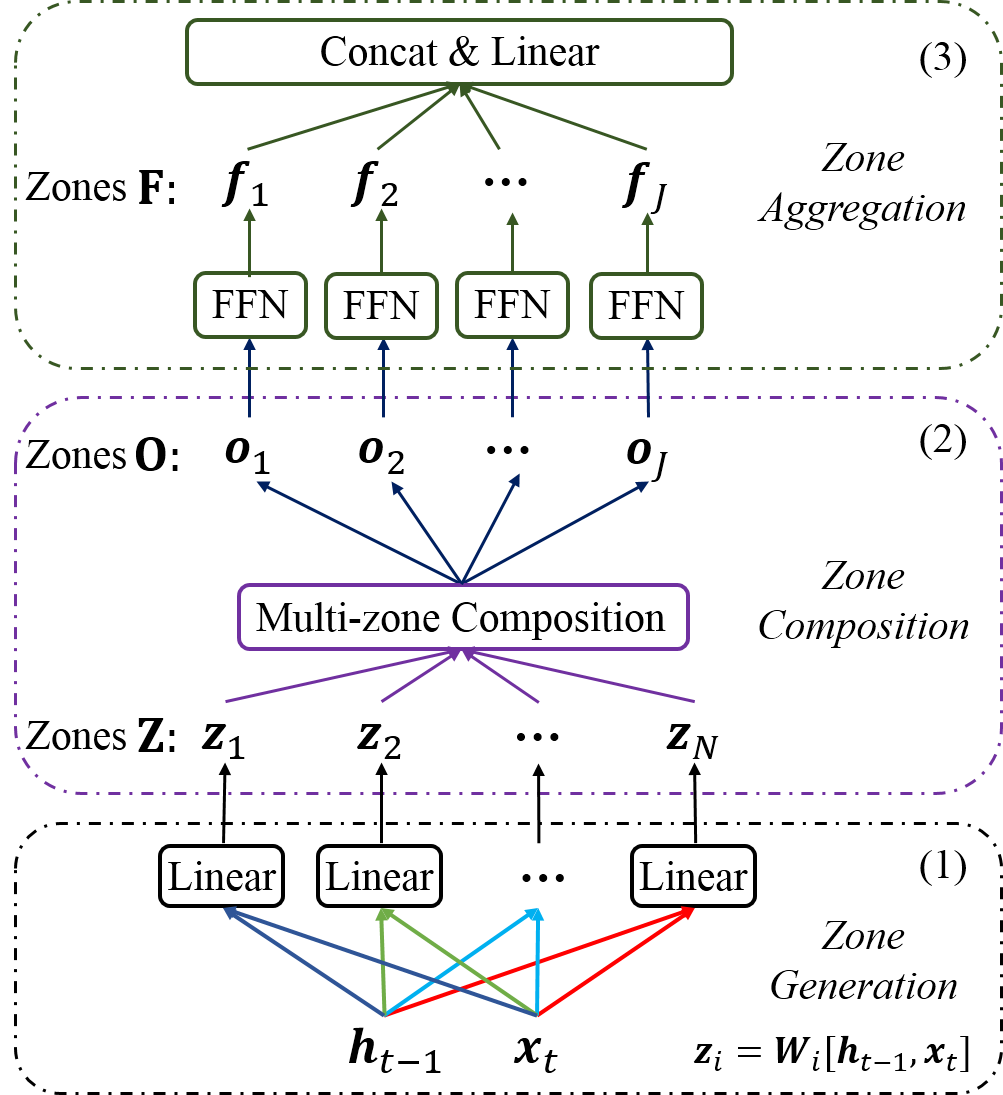}
      \caption{Illustration for the multi-zone transformation function, which is consists of three components. (1) \emph{Zone Generation}: linearly project [$\mathbf{x}_t, \mathbf{h}_{t-1}$] into $N$ zones $\mathbf{Z}$; (2) \emph{Zone Composition}: conduct  compositions on $\mathbf{Z}$ and output $J$ zones $\mathbf{O}$; (3) \emph{Zone Aggregation}: conduct a deep abstraction on $\mathbf{O}$ with a position-wise FFN and output new zones $\mathbf{F}$, then aggregate the representations by concatenating $\mathbf{F}$ and finally conduct a linear transformation to obtain the final representation.
   }  
   \label{f:mz_function} 
 \end{center}
\end{figure}

\subsection{Multi-Zone Transformation Function} \label{mzf}
The multi-zone transformation function, i.e., the $M$-function (in Eq.~(\ref{activate})~(\ref{gate})) is designed for enhancing both the transformation module (i.e., the candidate activation $\mathbf{\widetilde{h}}_{t}$) and the gating module (i.e., the $\mathbf{g}_t$). The $M$-function only takes the embedding $\mathbf{x}_t$ and the previous hidden state $\mathbf{h}_{t-1}$ as inputs, where $\mathbf{x}_t \in \mathbb{R}^{d_x}$ and $\mathbf{h}_{t-1} \in \mathbb{R}^{d_h}$. Figure~\ref{f:mz_function} shows the framework of the $M$-function, which consists of three components: \emph{Zone Generation}, \emph{Zone Composition}, and \emph{Zone Aggregation}. 
\paragraph{Zone Generation.}
As in Figure~\ref{f:mz_function}, we linearly project the current input and the previous hidden state into $N$ zone vectors $\mathbf{Z}=\{\mathbf{z}_1, \mathbf{z}_2, \dots, \mathbf{z}_N\}$ parameterized with different weight matrices. Each zone $\mathbf{z}_i$ is computed as:
\begin{eqnarray}
\mathbf{\mathbf{z}}_{i} = \mathbf{W}_{i}[\mathbf{x}_t, \mathbf{h}_{t-1}]
\end{eqnarray}
where $\mathbf{W}_{i} \in \mathbb{R}^{(d_x+d_h) \times d_z}$ (we set $d_z=d_h/N$). 
\paragraph{Zone Composition.}
Then we conduct a multi-zone composition with fully connected interactions on $\mathbf{Z}$, and output $J$ zones $\mathbf{O}=\{\mathbf{o}_1, \mathbf{o}_2, \dots, \mathbf{o}_J\}$, where $J$ can be equal to $N$ or not. In this step, zones in $\mathbf{Z}$ can fully interact with each other and generate new representations through our specially designed composition models. This composition can be designed flexibly, in particular, we provide three models, i.e., a self-attention based model, a graph convolutional model and a capsule networks based model, which will be formally described later.
\paragraph{Zone Aggregation.}
We further conduct a deep abstraction on the newly generated zones $\mathbf{O}$, and aggregate them to the final representation in this step. In particular, we apply a position-wise feed-forward network (FFN)~\cite{VaswaniEtal2017} to each zone $\mathbf{o}_j$ to conduct a deep abstraction and feature extraction, and output to $\mathbf{f}_j$ correspondingly. This consists of two linear transformations with a ReLU activation in between:
\begin{equation}
\mathbf{f}_j = \max(0, \mathbf{o}_j \mathbf{W}_1+\mathbf{b}_1)\mathbf{W}_2+\mathbf{b}_2 \label{ffn}
\end{equation}
where $\mathbf{W}_1 \in \mathbb{R}^{d_o\times d_f}$, and $\mathbf{W}_2 \in \mathbb{R}^{d_f\times d_o}$, $d_f$ is the filter size.
The linear transformations are the same across different zones. After that, we obtain $J$ deeply transformed zones $\mathbf{F}=[\mathbf{f}_1,\mathbf{f}_2,...,\mathbf{f}_J]^{\top}$. Then, we aggregate the representations by concatenating these zones and conduct a linear transformation to obtain the final representation.

\subsection{Models for Zone Composition} \label{mz_models}
To be noticed that, how to conduct full interactions on the multiple zones $\mathbf{Z}$ and generate new zones $\mathbf{O}$, i.e., $\mathbf{Z}\to\mathbf{O}$ in the ``\emph{Zone Composition}" (Figure~\ref{f:mz_function}-(2)) is crucial. It can be flexibly designed, in this paper, we just propose three effective approaches:

1) We apply self-attention~\cite{VaswaniEtal2017} to compute the relevance between each zone pair in $\mathbf{Z}$, and generate new zones $\mathbf{O}$ by a weighted sum of linear transformed  $\mathbf{Z}$, since self-attention has been proved effective to draw global dependencies of representations on different positions.

2) We exploit Graph Convolutional Network (GCN)~\cite{kipf2016variational,kipf2016semi} to conduct interactions on zones in $\mathbf{Z}$ by taking the zones as nodes of a graph, and generate new zones $\mathbf{O}$ by absorbing information from their neighbor nodes along weighted edges. As a special form of Laplacian Smoothing~\cite{Li2018DeeperII}, GCN takes the graph connectivity structure as a filter to perform neighborhood mixing, and therefore better for zone composition.

3) We utilize the capsule networks~\cite{hinton2011transforming,sabour2017dynamic,hinton2018matrix}, which have strong capabilities in representation composition and have been successfully applied to some NLP tasks~\cite{Zhaoemnlp2018,WangQi2018,Zhangaaai2019}. In particular, we model part-whole relationships between the older zones $\mathbf{Z}$ and the new zones $\mathbf{O}$ to aggregate diverse and useful information from older ones $\mathbf{Z}$, and achieve better representation composition through iterations of dynamic routing~\cite{sabour2017dynamic}.

\subsubsection*{Self-attention based MZU ($sat$MZU)}\label{sat}
We firstly project the $N$ zones into $N$ queries, $N$ keys and $N$ values of dimension $d_z$, each with a linear transformation, parameterized with different weights. 
Then we compute the dot products of the query with all keys, divide each by $\sqrt{d_z}$, and apply a $\mathrm{softmax}$ function to obtain the weights on the values. In practice, we compute the attention function on a set of queries simultaneously, packed together into a matrix $\mathbf{Q}$. The keys and values are also packed together into matrices $\mathbf{K}$ and $\mathbf{O}$. Then the matrix of outputs is computed as:
\begin{eqnarray}
\mathrm{Attention}(\mathbf{Q},\mathbf{K},\mathbf{O}) = \mathrm{softmax}\Bigg(\frac{\mathbf{Q}\mathbf{K}^\top}{\sqrt{d_z}}\Bigg)\mathbf{O}, \label{self-attention}
\end{eqnarray}
which is the $N$ new zones $\mathbf{O}=[\mathbf{o}_1,\mathbf{o}_2,...,\mathbf{o}_N]^{\top}$.

\subsubsection*{Graph Convolutional MZU ($gcn$MZU)} \label{gcn}
We take the $N$ zones as $N$ nodes and construct a graph $G=(V, E)$, where $V$ and $E$ represent the sets of nodes and edges, with $|V|=N$ and $|E|=N*N$, respectively.
Given node $\mathbf{z}_i$ and node $\mathbf{z}_j$, we use $\mathrm{cosine}$ to compute the semantic relevance between two nodes, as the weighted edge $w_{ij}$, which is calculated by:
\begin{equation}
{w_{ij}} = \frac{{\mathbf{z}_i}^{\top} \mathbf{z}_j} {{\left\| {\mathbf{z}_i} \right\|}\cdot {\left\| {\mathbf{z}_j} \right\|}}
\label{squash}
\end{equation}
where ${\left\| \cdot \right\|}$ stands for the L2 norm. Then we obtain the adjacent matrix\footnote{Self-connections have been added after the computation.} $\tilde{A} \in \mathbb{R}^{|V|\times|V|}$:
\begin{equation}
\tilde{A}=\left[
  \begin{array}{cccc}
    w_{11} &w_{12}  &... &w_{1N} \\
     ...&...  &... &...  \\
    w_{N1} &w_{N2} &...&w_{NN}\\
  \end{array}
\right]
\end{equation}
The degree matrix is represented as $\tilde{D}$, where $\tilde{D}_{ii} = \sum_j\tilde{A}_{ij}$. Meanwhile, we concatenate the representation of each node to a matrix $\mathbf{Z}={[\mathbf{z}_1,\mathbf{z}_2,...,\mathbf{z}_N]}^{\top} \in \mathbb{R}^{|V|\times d_z}$. Then we obtain the new node matrix $\mathbf{O}$ through a layer of convolution, which has incorporated the structured interactive information, as follows:
\begin{equation}
\mathbf{O} = \sigma \Big(\tilde{D}^{-\frac{1}2}\tilde{A}\tilde{D}^{-\frac{1}2}\mathbf{Z}\mathbf{W}_{g}\Big)
\end{equation}
where $\mathbf{W}_{g}\in \mathbb{R}^{|V|\times d_g}$ is the weight matrix, and $d_g$ (we set $d_g=d_z$) is the dimension of this convolutional layer, and $\mathbf{O}\in \mathbb{R}^{|V|\times d_g}$. It is a special form of Laplacian Smoothing~\cite{Li2018DeeperII}, which computes the new representation of a node as the weighted average of itself and its neighbors'.

\subsubsection*{Capsule Networks based MZU ($cap$MZU)} \label{cap}
We take the $N$ zones $\mathbf{Z}=\{\mathbf{z}_1, \mathbf{z}_2, \dots, \mathbf{z}_N\}$ as $N$ low-level capsules, and aim to output $J$ high-evel capsules (i.e., zones) $\mathbf{O}=\{\mathbf{o}_1, \dots, \mathbf{o}_J\}$, where $\mathbf{o}_j \in \mathbb{R}^{d_o}$ (we set $d_o=d_h/J$). We aggregate information from $\mathbf{Z}$ and conduct representation composition to generate $\mathbf{O}$, by modeling part-whole relationships between capsules in $\mathbf{Z}$ and capsules in $\mathbf{O}$.

We firstly initialize the initial logits $b_{ij}$ with 0, which are log prior probabilities that capsule $\mathbf{z}_i$ should be coupled to capsule $\mathbf{o}_j$. And we generate the ``prediction vectors" ${\bf \hat{z}}_{j|i}$ from the low-level capsules as follows:
\begin{equation}
{\bf \hat{z}}_{j|i} = {\bf z}_i{\bf W}^c_{ij} \label{z_ij}
\end{equation}
where ${\bf W}^c \in \mathbb{R}^{N\times J \times d_z \times d_o}$ is the weight tensor.\footnote{In practice, we share the transformation matrix of each output capsule $o_j$ among all the input capsules, that is ${\bf W}^c \in \mathbb{R}^{J \times d_z \times d_o}$.}

Then we aggregate information from $\mathbf{Z}$ and conduct representation composition to generate $\mathbf{O}$, through $T$ iterations of dynamic routing on ${\bf \hat{z}}_{j|i}$. At each iteration, we firstly compute coupling coefficient $c_{ij}$ between low-level capsule $\mathbf{z}_i$ and all high-level capsules, by a ``softmax'' on initial logits $b_{ij}$, as follows:
\begin{equation}
c_{ij} = \mathrm{softmax}({\bf b}_i) = \frac{\exp(b_{ij})}{\sum_k \exp(b_{ik})} \label{softmax}
\end{equation}
After that we generate ${\bf s}_j$ by a weighted sum over ${\bf \hat{z}}_{j|i}$ with coupling coefficient $c_{ij}$ as weights, computed as:
\begin{equation}
{\bf s}_j = \sum_i{c_{ij}{\bf \hat{z}}_{j|i}}  \label{s_j}
\end{equation}
Then we take ${\bf s}_j$ as input and output capsule $\mathbf{o}_j$ by squashing ${\bf s}_j$ as follows:
\begin{equation}
{\bf o}_j = \mathrm{squash}({\bf s}_j) = \frac{||{\bf s}_j||^2}{1+||{\bf s}_j||^2} \frac{{\bf s}_j}{||{\bf s}_j||}
\label{squash}
\end{equation}
where $\mathrm{squash}(\cdot)$ ~\cite{sabour2017dynamic} is a non-linear function to ensure that short vectors get shrunk to almost zero length and long vectors get shrunk to a length slightly below $1$. 

Then $b_{ij}$ are iteratively refined by measuring the agreement (scalar product) between the current output ${\bf o}_j$ and the prediction ${\bf \hat{z}}_{j|i}$:
\begin{equation}
b_{ij} = b_{ij} + {\bf \hat{z}}_{j|i} \cdot {\bf o}_j \label{b_ij}
\end{equation}
After $T$ iterations of dynamic routing, we obtain $J$ final capsules (zones) $\mathbf{O}=[\mathbf{o}_1,\mathbf{o}_2,...,\mathbf{o}_J]^{\top}$.

\subsection{Deep Transition MZU}
The MZU is generic as standard RNNs, and can be extended to any deep MZUs. Since recent studies~\cite{pascanu2014construct,barone2017deep,meng2018dtmt} have demonstrated the superiority of deep transition RNNs over deep stacked RNNs, we extend the MZU to a more powerful deep transition MZU. The entire deep transition block is composed of one MZU cell followed by several {\bf t}ransition MZU ($t$-MZU) cells at each time step. As a special case of MZU, $t$-MZU only has ``state" as input, i.e. the embedding input of $t$-MZU is a zero vector. In the whole recurrent procedure, for the current time step, the ``state" output of one MZU/$t$-MZU cell is used as the ``state" input of the next $t$-MZU cell. And the ``state" output of the last $t$-MZU cell for the current time step is carried over as the ``state" input of the first MZU cell for the next time step. 

For a $t$-MZU cell, each hidden state at time step $t$ at transition depth $l$ is computed as follows:
\begin{eqnarray}
\mathbf{h}_t^{l} &=& (1 - \mathbf{g}_t^{l}) \odot \mathbf{h}_t^{l-1} + \mathbf{g}_t^{l} \odot \mathbf{\widetilde{h}}_t^{l} \\
\mathbf{\widetilde{h}}_t^{l} &=& \tanh(M_{h}^{l}(0, \mathbf{h}_t^{l-1})) \label{t-activate}
\end{eqnarray}
where the gate $\mathbf{g}_t^{l}$ is computed as:
\begin{eqnarray}
\mathbf{\mathbf{g}}_t^{l} = \sigma(M_{g}^{l}(0, \mathbf{h}_t^{l-1})) \label{t-gate}
\end{eqnarray}
where $M_{h}^{l}$ and $M_{g}^{l}$ are Multi-Zone Transformation functions, as described in Section~\ref{mzf}.

\subsection{Regularization on Multiple Zones}
We expect that the representations in multiple zones of MZU are as different as possible. Thus we propose a disagreement regularization on these zones, inspired by~\citeauthor{li2018multi}~\shortcite{li2018multi}.
This regularization is designed to maximize the cosine distance (i.e., negative cosine similarity) between zone pairs.
Our objective is to enlarge the average cosine distance among all zone pairs, computed as:
\begin{equation}\label{eq:sub}
D_\mathrm{zone} = - \frac{1}{N^2} \sum_{i=1}^{N}\sum_{j=1}^{N} \frac{{\mathbf{z}_i}^{\top} \mathbf{z}_j} {{\left\| {\mathbf{z}_i} \right\|}\cdot {\left\| {\mathbf{z}_j} \right\|}} \label{zone}
\end{equation}

\paragraph{Training Objective.}
Taking into the regularization, the training objective of MZU is:
\begin{equation}
J(\theta) = \mathop{\arg\max}_{\theta}\Bigg\{\underbrace{L(\theta)}_{\mathrm{likelihood}} + \lambda*\underbrace{\sum^{s}\sum^{m} D_\mathrm{zone}(\theta)}_{\mathrm{disagreement}} \Bigg\} \label{final_objective}
\end{equation}
where $m$ is the number of $M$-functions in the MZU, $s$ is the number of inputs (e.g., tokens), $L(\theta)$ is a task specific training objective, and $\lambda$ is a hyper-parameter used to balance the preference between two terms of the loss function.

\section{Experiments}
We verify the MZU on the character-level language modeling task and the aspect based sentiment analysis task. 

\subsection{Character-level Language Modeling}\label{lm}
We use two standard datasets, namely the Penn Treebank~\cite{marcus1993building} and the larger Wikipedia dataset (text8)~\cite{mahoney2011large}.

{\bf Penn Treebank.} The Penn Treebank dataset is a collection of Wall Street Journal articles written in English. 
We follow the process procedure introduced in~\cite{mikolov2012subword}, and split the data into training, validation and test sets consisting of 5.0M, 390K and 440K characters, respectively.

{\bf The Wikipedia Corpus (text8).} The text8 dataset consists of 100M characters extracted from the English Wikipedia. text8 contains only alphabets and spaces, and thus we have total 27 symbols. In order to compare with other previous works, we follow the the process procedure in~\cite{mikolov2012subword}, and split the data into training, validation and test sets consisting of 90M, 5M, and 5M characters, respectively.

\paragraph{Training Details.}
We train the model using Adam~\cite{kingma2014adam} with an initial learning rate of 0.001. Each update is done by using a mini-batch of 256 examples. We use the truncated backpropagation through time to approximate the gradients, setting the length to 150. The norm of the gradient is clipped with 5.0. We apply layer normalization to our models, and apply dropout to the candidate activation to avoid overfitting. The dropout rate for the Penn Treebank task and the text8 task are set to 0.5 and 0.3, respectively. The hidden size and filter size for Penn Treebank are set to 800 and 1000, respectively. And those for text8 are set to 1,536 and 3,072, respectively. The embedding size is 256. We share parameters of the $M$-function across different depths (i.e., the deep transition counterpart) for Penn Treebank. We measure the performance of checkpoints by evaluating \emph{bits per character} (BPC) over the valid set, and make the test with the best one. BPC is the negative log-likelihood divided by natural logarithm of 2. The lower the better. 

\paragraph{System Description.}
We use $sat$MZU, $gcn$MZU and $cap$MZU standing for the self-attention based MZU, the graph convolutional MZU and the capsule networks based MZU, respectively. ``DT" stands for deep transition MZU cells, with transition depth setting to 1 by default, i.e. one MZU cell plus one $t$-MZU cell. 
For all MZU models, we apply 4 zones in $M$-functions, except when otherwise mentioned. For all $cap$MZUs, we set the number of output capsules to 2, and use 3 routing iterations, according to primary experiments. We set $\lambda$ to 1.0 in Eq.~\ref{final_objective} for all experiments. From primary experiments with different settings to $\lambda$, we find setting $\lambda>0$ (promoting to different zones) consistently improve the performance (0.005 $\sim$ 0.007 BPC) on both shallow and deep MZU models. While setting $\lambda<0$ (promoting to similar zones), e.g. -0.5 or -1.0, sharply declines the performance (0.03 $\sim$ 0.05 BPC). This indicates modeling multi-zone transformation affects the state transition, and consequently affects the performance. 

\begin{table}[t!]
\centering
\scalebox{0.92} {
\begin{tabular}{l| r| l}
\hline
\bf \textsc{Model} & \bf \textsc{Size} & \bf \textsc{BPC}  \\
\hline
LSTM~\small{\cite{krueger2016zoneout}}              & --    & 1.36 \\
Zoneout LSTM~\small{\cite{krueger2016zoneout}} 	    & --    & 1.27 \\
2-Layers LSTM~\small{\cite{zoph2016neural}}         &6.6M 	& 1.243 \\
LN-HM-LSTM~\small{\cite{chung2017hierarchical}}    	& --    & 1.24 \\
HyperLSTM~\small{\cite{ha2016hypernetworks}}        &14.4M  & 1.219 \\
NASCell~\small{\cite{zoph2016neural}}               &16.3M  & 1.214 \\
FS-LSTM-4~\small{\cite{mujika2017fast}}             &7.2M 	& 1.190 \\
res-IndRNN~\small{\cite{li2018independently}}       & -- & 1.19 \\
\hline
$sat$MZU+DT (ours) & 6.5M 	& 1.189 \\
$gcn$MZU+DT (ours) & 6.4M   & 1.183 \\
$cap$MZU+DT (ours) & 7.2M 	& \bf{1.181} \\
\hline
\end{tabular}
}
\caption{BPC on Penn Treebank test set.}  
\label{t:result-ptb-char}
\end{table}

\begin{table}[t!]
\centering
\scalebox{0.92} {
\begin{tabular}{l| r| l}
\hline
\bf \textsc{Model} & \bf \textsc{Size} & \bf \textsc{BPC}  \\
\hline
Large td-LSTM~\small{\cite{zhang2016architectural}}  & --  & 1.49 \\
MI-LSTM~\small{\cite{wu2016multiplicative}}          & --  & 1.44 \\
BN-LSTM~\small{\cite{cooijmans2016recurrent}}        & --  & 1.36 \\
Zoneout LSTM~\small{\cite{krueger2016zoneout}} 	     & --  & 1.336 \\
LN HM-LSTM~\small{\cite{chung2017hierarchical}}      & 35M & 1.29 \\
Large RHN~\small{\cite{zilly2017recurrent}}          & 45M & 1.27 \\
Large mLSTM~\small{\cite{krause2016multiplicative}}  & 45M & 1.27 \\
\hline
$sat$MZU+DT (ours) & 32M & 1.268 \\
$gcn$MZU+DT (ours) & 32M & 1.260 \\
$cap$MZU+DT (ours) & 41M & {\bf 1.249} \\
\hline
Transformer~\small{\cite{al2018character}}       & 235M  &  1.13 \\
Transformer-XL~\small{\cite{dai2019transformer}} & 277M  &  1.08 \\
\hline
\end{tabular}
}
\caption{BPC on text8 test set.} 
\label{t:result-text8-char}
\end{table}

\paragraph{Main Results.}
Table~\ref{t:result-ptb-char} shows results on Penn Treebank test set, with comparing to existing competitive approaches. Our MZU models outperform previous competitive approaches, including deeper RNNs, e.g., the 4-layers FS-LSTM and the 11-layers res-IndRNN, without using more parameters. Among three MZU models, the $gcn$MZU+DT and the $cap$MZU+DT achieve better performance than the $sat$MZU+DT, and the $cap$MZU+DT achieves the best BPC with 1.181. The $cap$MZU can model part-whole relationships between the old zones and the new zones to aggregate useful information, through multiple iterations of routing-by-agreement. Such kind of multi-round refinements let the new zones contain more accurate representations and therefore $cap$MZU preforms better than other MZUs.

To demonstrate that MZUs work well across datasets, we conduct experiments on a larger dataset text8. As shown in Table~\ref{t:result-text8-char}, using even less parameters, our $sat$MZU+DT, $gcn$MZU+DT and $cap$MZU+DT outperform several previous deep RNNs, including the 10-depths Large RHN and the stacked Large mLSTM, except the recently proposed transformer-based models~\cite{al2018character,dai2019transformer} trained with much longer context (512), deeper layers (24-layers Transformer \& 64-layers Transformer-XL) and larger amount of parameters. Explicit using longer context indeed improves the performance of language modeling. While in this work, we mainly focus on boosting the transition functions (i.e., only operate on $\mathbf{x}_t$ and $\mathbf{h}_{t-1}$) of RNNs. Our approaches can also be applied on transformer-based architectures, e.g for enhancing the multi-head attention.

\begin{table}[t!]
\centering
\scalebox{0.98} {
\begin{tabular}{c| c| c | c}
\hline
\bf \textsc{\#Zones} & $sat$MZU  & $gcn$MZU  & $cap$MZU \\
\hline
2 & 1.191  & 1.187 & 1.186\\
4 & {\bf 1.189} & {\bf 1.183} & \bf{1.181}\\
8 & 1.200 & 1.189  & \bf{1.181}\\
16 & 1.208 & 1.185 & 1.186\\
\hline
\end{tabular}
}
\caption{Effects of zone numbers on Penn Treebank test set. All the tested models are with ``DT", which omitted due to the limited table space.}   
\label{t:number-subspace}
\end{table}

\begin{table}[t!]
\centering
\scalebox{0.98} {
\begin{tabular}{l| c| c}
\hline
\bf \textsc{Model}  & \bf \textsc{No-DT} & \bf \textsc{DT} \\
\hline
GRU                          &1.269 & 1.208 \\
\hline
$sat$MZRNN                   &{\bf 1.214} & {\bf 1.189} \\
~~~~~\textsc{Regular-Gate}   &1.236 & 1.207 \\
~~~~~\textsc{Regular-Trans.} &1.247 & 1.208 \\
\hline
$gcn$MZRNN                   &{\bf 1.196} & {\bf 1.183} \\
~~~~~\textsc{Regular-Gate}   &1.224 & 1.197 \\
~~~~~\textsc{Regular-Trans.} &1.225 & 1.191 \\
\hline
$cap$MZRNN                   &{\bf 1.192} & {\bf 1.181} \\
~~~~~\textsc{Regular-Gate}   &1.219 & 1.190 \\
~~~~~\textsc{Regular-Trans.} &1.211 & 1.188 \\
\hline
\end{tabular}
}
\caption{Ablation study of $M$-functions in MZUs. ``\textsc{Regular}" stands for replacing the $M$-function with the regular linear transformation function on $[\mathbf{x}_t, \mathbf{h}_{t-1}]$.} 
\label{t:ablation}
\end{table}

\paragraph{The Number of Zones.}
We investigate the impact of zone numbers on MZUs, in particular, we conduct experiments with three deep MZU models (MZU+DT) on Penn Treebank. For each model, we respectively test 2, 4, 8, and 16 zones in $M$-functions. As the dimension of the zone is decided by the hidden size and the zone number, i.e., $d_z$ = $d_h / N$, more zones do not lead to more parameters. As shown in Table~\ref{t:number-subspace}, using 4 zones achieves the best performance for all MZU models. Additionally, we observe that the $sat$MZU is more sensitive on the number of zones, and leveraging too many zones (e.g. 8, 16) sharply declines the performance. The $gcn$MZU and the $cap$MZU are more robust on modeling multiple zones, benefiting from the powerful graph convolutional structure and dynamic routing algorithm, respectively. 

\begin{figure}[t!]
\begin{center}
      \includegraphics[width=0.23\textwidth]{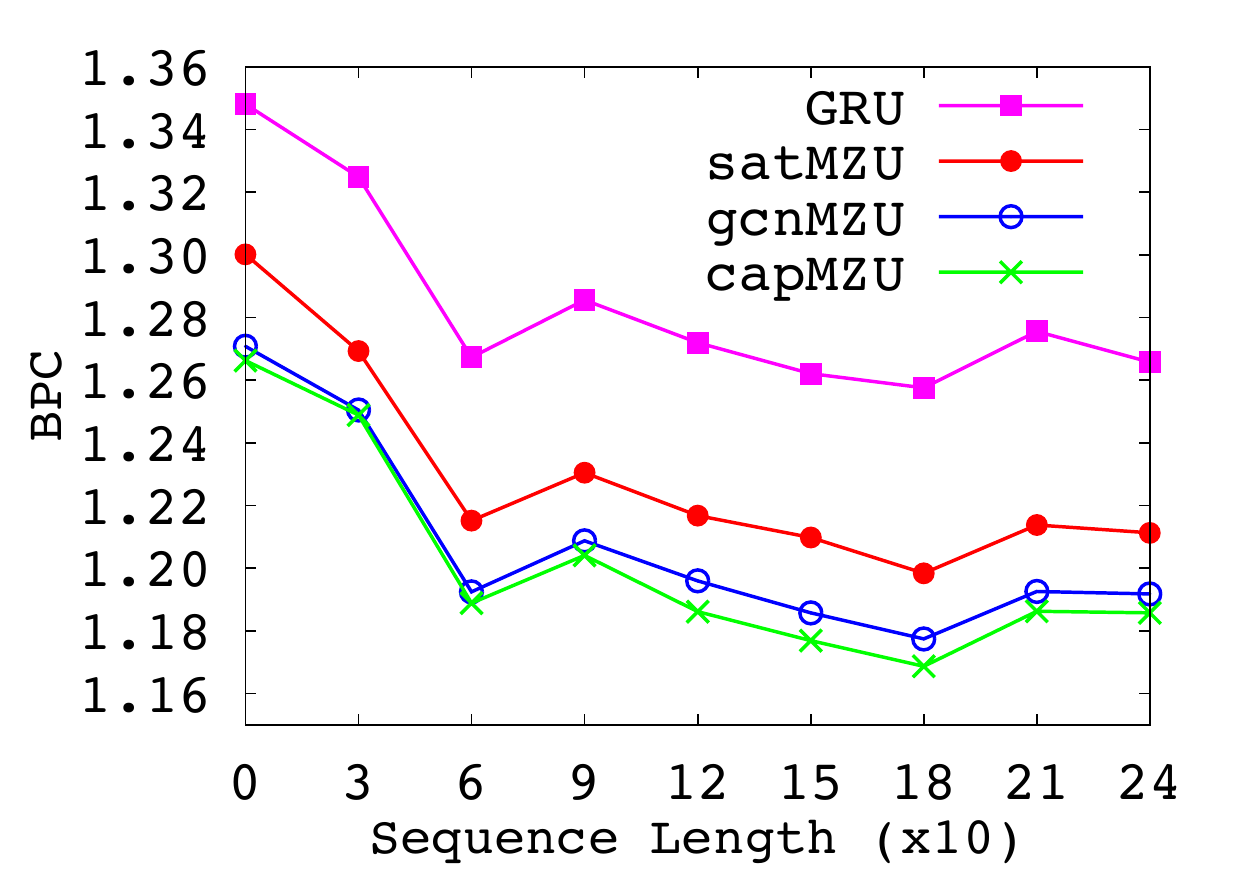}
      \hspace{-6pt}
      \includegraphics[width=0.23\textwidth]{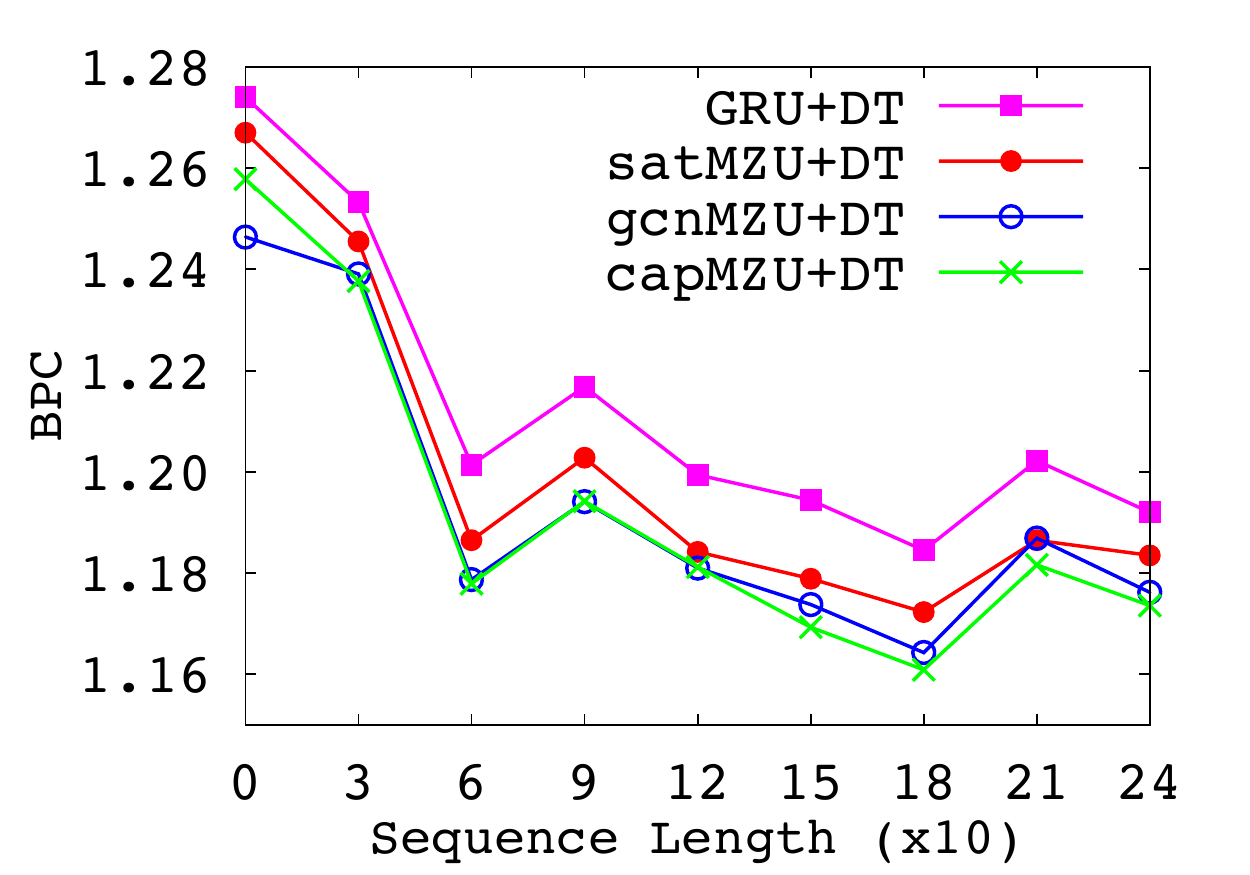}
      \caption{BPC of shallow (left) and deep (right) models on sequences with different lengths. The numbers on X-axis stand for sequences longer than the corresponding length (x10), e.g., 9 for sequences with (90, 120] characters. The superiority of MZUs over GRUs is more obvious on long sequences (e.g. $>$90).
      } \label{f:length_base}
 \end{center} 
\end{figure}

\paragraph{Ablation Study.}
Since the $M$-function works on the gating module (i.e., $\mathbf{g}_t$) and the transformation module (i.e., $\mathbf{\widetilde{h}}_{t}$), we conduct an ablation study to investigate its effects on both modules. We show results in Table~\ref{t:ablation}, where ``\textsc{Regular}" stands for replacing the $M$-function with the regular linear transformation on $[\mathbf{x}_t, \mathbf{h}_{t-1}]$. As shown in Table~\ref{t:ablation}, the $M$-function is crucial for both the gating module (``\textsc{Gate}") and the transformation module (``\textsc{Trans.}"), for both shallow (``\textsc{No-DT}") and deep (``\textsc{DT}") MZU models, since replacing any module with the \textsc{Regular} one sharply declines the performance. We also list results\footnote{We also apply dropout and layer normalization to GRUs, and build models with the same settings to our MZUs.} of GRU and deep transition GRU~\cite{pascanu2014construct} (i.e., ``\textsc{DT}") in Table~\ref{t:ablation}. Even with the module degeneration, our MZUs still outperform GRUs which only based on single space composition, in shallow and deep models, correspondingly.

\paragraph{About Length.}
We investigate effects of MZU models on sequences with different length, and compare them with GRU and deep transition GRU (GRU+DT) in Figure~\ref{f:length_base}. The results of shallow and deep models are in the left and right subfigure, respectively. As shown in Figure~\ref{f:length_base}, on sequences with different length, shallow and deep MZU models consistently yield better performance than the GRU and the GRU+DT, respectively. And the superiority is more obvious on long sequences (e.g., $>$90) for both shallow and deep MZU models, demonstrating the stronger capability for long sequence processing of the MZU.

\begin{figure}[t!]
\begin{center}
      \includegraphics[width=0.48\textwidth]{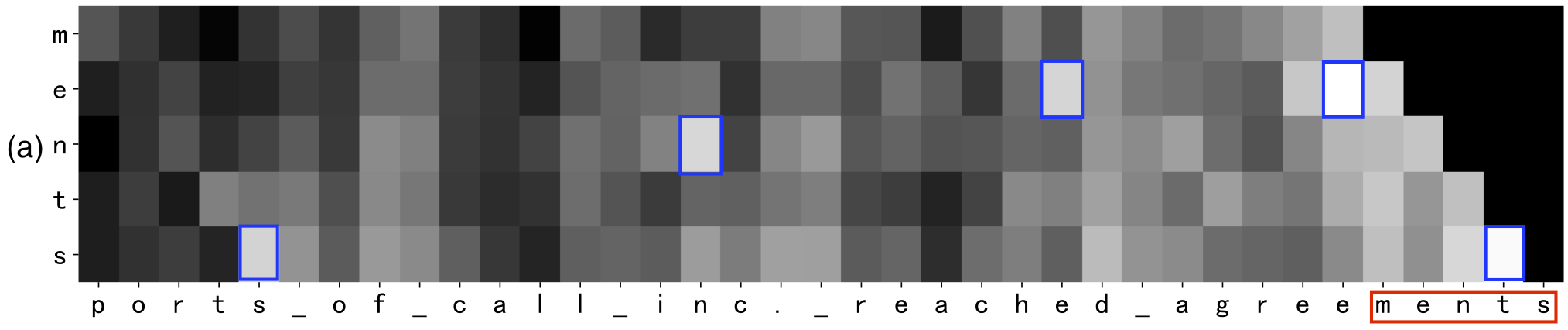}
      \includegraphics[width=0.48\textwidth]{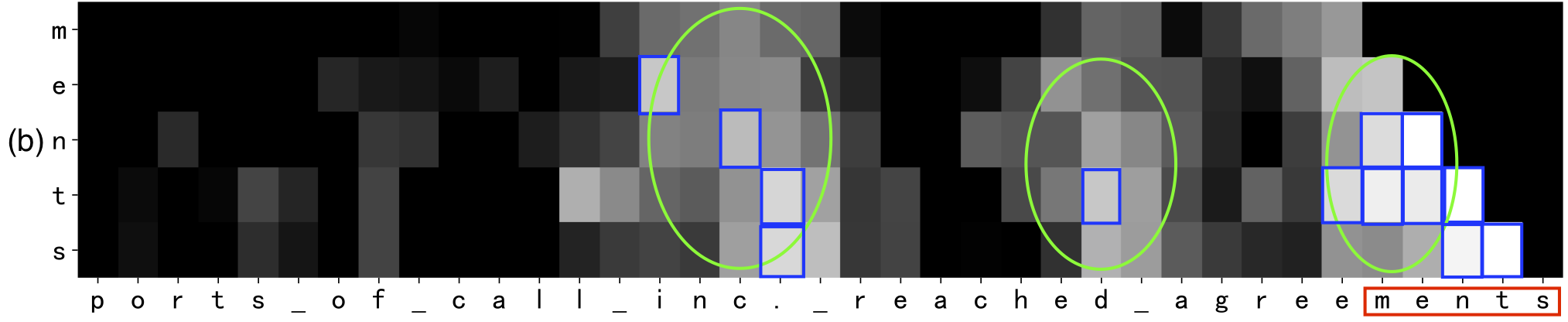}
      \includegraphics[width=0.48\textwidth]{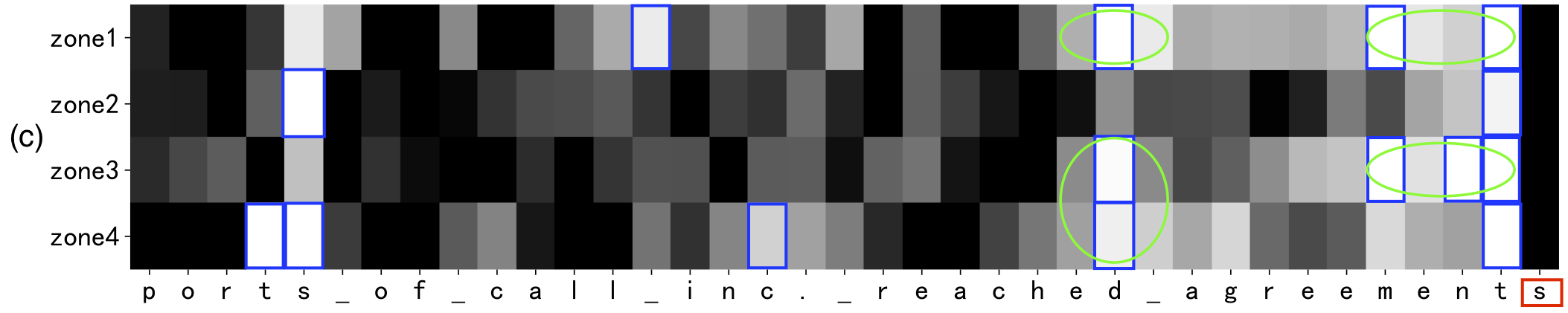}
      \caption{Relevance results between each of the last five characters (i.e., `m', `e', 'n', `t', `s') and all characters before it, from the GRU(a) and the $cap$MZU(b). (c) shows the zone-specific relevance result of $cap$MZU between the last character `s' and all characters before it. The higher the relevance, the more white of the grayscale. Blue rectangles and green ellipses indicate high-relevance parts. `\_' stands for the space between two words.} \label{f:case_study}   
 \end{center}
\end{figure}

\paragraph{Visualization Analysis.}
We conduct a visualization analysis to demonstrate the superiority of the MZU (Figure~\ref{f:case_study}(b)) over GRU (Figure~\ref{f:case_study}(a)).
In particular, we compare the $cap$MZU to GRU. We compute the relevance between each of the last five characters (i.e., `m', `e', 'n', `t', `s') and all characters before it. For example, the relevance between `m' and the first character `p' is calculated by the $cosine$ similarity between the candidate activation (stands for the transformation at current step) of `m' and the hidden state of `p'. The higher the relevance, the more white of the grayscale. From Figure~\ref{f:case_study}(a) (GRU), we can find that 1) The gray distribution is not sharply contoured; and 2) the highest relevance positions to each computed character usually occur at the nearest position to it or at positions with the same surface character, such as the boxes marked blue. In Figure~\ref{f:case_study}(b) ($cap$MZU), the gray distribution is more sharply contoured, e.g., three regions marked by green ellipses. Additionally, it captures some philological segments, such as the suffixes ``ments" and ``ed", and parts of the word ``inc.", which are related to the word ``agreements". Furthermore, we investigate how multiple zones affect the performance, and show the zone-specific relevance between the last character `s' and all characters before it in Figure~\ref{f:case_study}(c) ($cap$MZU). We find that different zones indeed capture different philological information. For example, ``zone1" and ``zone3" capture some suffixes (e.g., ``ment", ``ed''), ``zone2" captures the same surface character `s' in long distance, and ``zone4" not only captures the nearest `t' to `s', but also captures the long distance and frequent collocation ``ts".

\begin{table}[t]
\centering
\scalebox{0.9} {
\begin{tabular}{c|c|c}
\hline 
\textbf{\textsc{Sentence}}  &\multicolumn{2}{l} {\makecell{The appetizers are ok, but the service is slow. }}  \\ \hline
\textbf{Aspect}    & \qquad The appetizers \quad   \qquad & \qquad service \\ \hline
\textbf{Sentiment} & \qquad Neutral \quad \qquad & \qquad Negative \\
\hline
\end{tabular}
} 
\caption{An example that contains different sentiment polarities towards two aspects.} 
\label{tbl:testE}
\end{table}

\begin{table*}[t!]
\centering
\scalebox{0.95} {
\begin{tabular}{l|ll|ll}
\hline
\multirow{2}{*}{\bf \textsc {Models}}    & \multicolumn{2}{c|}{\bf \textsc{Restaurant}}   & \multicolumn{2}{c}{\bf \textsc{Laptop}}             \\ \cline{2-5}
& \multicolumn{1}{c}{DS} & \multicolumn{1}{c|}{HDS} & \multicolumn{1}{c}{DS} & \multicolumn{1}{c}{HDS}     \\ \hline
TD-LSTM*~\cite{Tang:16a}    & 73.44$\pm$1.17           & 56.48$\pm$2.46            & 62.23$\pm$0.92            & 46.11$\pm$1.89   \\
ATAE-LSTM*~\cite{Wang:16}   & 73.74$\pm$3.01           & 50.98$\pm$2.27            & 64.38$\pm$4.52            & 40.39$\pm$1.30   \\
GCAE*~\cite{weixueGCAE:18}  & 77.28$\pm$0.32           & 56.73$\pm$0.56            & 69.14$\pm$0.32            & 47.06$\pm$2.45   \\ 
\hline
GRU (\textsc{With-DT})    & 78.25$\pm$0.51      & 56.17$\pm$0.72        & 71.01$\pm$0.73     & 47.03$\pm$1.15                 \\ 
$cap$MZRNN (\textsc{With-DT})    & \bf{79.40}$\pm$0.23      & \bf{57.59}$\pm$0.70        & \bf{72.45}$\pm$0.68     & \bf{47.96}$\pm$1.12      \\ 
\hline
\end{tabular}
}
\caption{The accuracy (mean of five times training with standard deviation) on SemEval 2014 term-based datasets. Results (marked with ``*") of existing systems are cited from~\cite{weixueGCAE:18}. } 
\label{tbl:result_atsa}
\end{table*}

\subsection{Aspect Based Sentiment Analysis} \label{absa}

\paragraph{Task Description.}
Since RNNs are widely used for sequence modeling and classification tasks, we choose the well-known and challenging ABSA task as a case study to demonstrate the generalizability across tasks of the MZU. We test on the aspect-term based task of the SemEval 2014 Task 4~\cite{Pontiki:14} with two datasets that contain domain-specific customer reviews for restaurants and laptops, each of which contains four sentiment labels (i.e., Positive, Negative, Neutral, and Conflict). For this task, the goal is to infer sentiment polarity over the aspect (i.e., the given term), which is a subsequence of the sentence. The example in Table~\ref{tbl:testE} shows the customer's different attitudes towards two terms: ``{\em The appetizers}'' and ``{\em service}''.
For each dataset, we follow~\cite{weixueGCAE:18,liangetal2019} and construct a hard sub-dataset (named ``HDS'') by extracting from the full dataset (named ``DS"). In ``HDS'', all sentences contain multiple aspects, each of which corresponds to a dfferent sentiment label. On this sub-dataset, we copy each sentence $n$ times, where $n$ is the number of aspects in a sentence.

\paragraph{Models.}
We compare our MZU with the deep transition GRU and several previous competitive systems. For the MZU, we use $cap$MZRNN with deep transition. We take the ABSA as a sentence-level classification task. Therefore, for both GRU model and $cap$MZRNN model, we follow~\cite{Tang:16a}, and firstly encode the sentence as a standard Bi-RNN encoder does. Then we conduct mean pooling on the output hidden states to generate a sentence vector. Next we concatenate the sentence vector with the given aspect embedding (the mean of embeddings for multi-word aspects). Finally, we predict the sentiment polarity with the concatenated representation by a softmax layer. 

\paragraph{Results.}
As shown in Table~\ref{tbl:result_atsa}, our implemented deep transition GRU outperforms most previous LSTM-based systems (i.e., TD-LSTM~\cite{Tang:16a}, ATAE-LSTM~\cite{Wang:16} and the gated CNN-based system GCAE~\cite{weixueGCAE:18} (comparable on ``HDS" sub-datasets), demonstrating that our deep transition GRU is a strong baseline. When incorporating the MZU, our $cap$MZRNN further improve the performance on multiple datasets by significant margins. These results demonstrate the superiority and generalizability of the MZU cross tasks.

\section{Related Work}
This work is inspired by the idea of processing attentions on multiple subspaces with different heads~\cite{VaswaniEtal2017}. While different from this multi-channel style operation, we conduct fully connected interactions on multiple zones to extract useful information. In particular, we propose three effective models to conduct interactions and representation composition.

Some researchers~\cite{pascanu2014construct,zilly2017recurrent,Graves16,mujika2017fast} boost the transition function with deep architectures, such as adding intermediate layers to increase the transition depth. Different from these works, we boost the performance of RNNs by modeling multiple zone composition for the transition function. Additionally, our MZU is generic and can be easily extended to deep MZUs. 

Some researchers devise models to capture specific features from the processed history sequence, including incorporating an attention mechanism into LSTM cells~\cite{ZhongLSTA2018} or extending the GRU with a recurrent attention unit~\cite{ZhongRAU2018}. Different from these works, we focus on boosting the transition function by modeling multiple zone composition, which only operates on $\mathbf{x}_t$ and $\mathbf{h}_{t-1}$, at each time step.

Some studies~\cite{merity2017regularizing,Siddhartha2018} investigate regularization techniques for optimizing LSTM-based models, such as weight dropout, variational dropout, weight tying, temporal activation regularization, and past decode regularization. These techniques can be used to further optimize MZU-based models in the future.

\section{Conclusion}
We propose a generic Multi-zone Unit for RNNs along with three effective variants (i.e., the $sat$MZU, the $gcn$MZU and the $cap$MZU), which boost the state transition function through modeling multiple space composition. To further enhance MZUs, we propose an effective regularization objective to promote the diversity in multiple zones. Experiments on two character-level language modeling datasets show that our MZUs can automatically capture rich linguistic information and substantially improve the performance. Experimental results on the ABSA task further demonstrate the superiority and generalizability of the MZU.

\section*{Acknowledgments}
Yang Liu is supported by the National Key R\&D Program of China (No. 2017YFB0202204), National Natural Science Foundation of China (No. 61761166008), Beijing Advanced Innovation Center for Language Resources (No. TYR17002). 

\bibliography{aaai}
\bibliographystyle{aaai}

\end{document}